\documentclass{article}
\usepackage{spconf,amsmath,epsfig}
\usepackage{algorithm}
\usepackage{xspace}
\usepackage{srcltx}
\usepackage{amsmath,epsfig, times}
\usepackage{amssymb,amsmath,graphicx,algorithm}
\usepackage[noend]{algpseudocode}

\ninept
\def\reg{{\rm\ooalign{\hfil
     \raise.07ex\hbox{\scriptsize R}\hfil\crcr\mathhexbox20D}}}

\ninept
\title{Accelerating Hessian-Free Optimization for Deep Neural \\ Networks by Implicit Preconditioning and Sampling}
\name{Tara N. Sainath, Lior Horesh, Brian Kingsbury, Aleksandr Y. Aravkin, Bhuvana Ramabhadran}
\address{IBM T. J. Watson Research Center,  Yorktown Heights, NY 10598, U.S.A\\
 \{tsainath, lhoresh, bedk, saravkin, bhuvana\}@us.ibm.com}

\begin{document}
\maketitle
Hessian-free training has become a popular parallel second order optimization technique for Deep Neural Network training. This study aims at speeding up Hessian-free training, both by means of decreasing the amount of data used for training, as well as through reduction of the number of Krylov subspace solver iterations used for implicit estimation of the Hessian. In this paper, we develop an L-BFGS based preconditioning scheme that avoids the need to access the Hessian explicitly. Since L-BFGS cannot be regarded as a fixed-point iteration, we further propose the employment of flexible Krylov subspace solvers that retain the desired theoretical convergence guarantees of their conventional counterparts. Second, we propose a new sampling algorithm, which geometrically increases the amount of data utilized for gradient and Krylov subspace iteration calculations. On a 50-hr English Broadcast News task, we find that these methodologies provide roughly a 1.5x speed-up, whereas, on a 300-hr Switchboard task, these techniques provide over a 2.3x speedup, with no loss in WER. These results suggest that even further speed-up is expected, as problems scale and complexity grows.
\section{Introduction}
Second order optimization techniques have been extensively explored for problems involving pathological curvature, such as deep neural network (DNN) training problems. In fact, \cite{Martens2010} demonstrated success of a second order technique, known as Hessian-free (HF) optimization \cite{ Horesh2006}, with DNNs on various image recognition tasks. In addition, \cite{bedk:hf} successfully applied the HF optimization technique with DNNs for speech recognition tasks. Other second order methods, including L-BFGS \cite{google:dist} and Krylov Subspace Descent \cite{povey:kyrlov}, have also shown great success for DNN training. 


Second order methods are particularly important for sequence-training of DNNs, which provides a 10-20\% relative improvement in WER over a cross-entropy (CE) trained DNN \cite{bedk:nn}. Because sequence training must use information from time-sequential lattices corresponding to utterances, sequence training is performed using utterance randomization rather than frame randomization. For mini-batch stochastic gradient descent (SGD), which is often used for CE training, frame randomization performs better than utterance randomization \cite{seide:sequence}. However, because sequence-training must be accomplished at the utterance level, second order methods perform much better than SGD, as second order methods compute the gradient over a large batch of utterances compared to utterance mini-batch SGD \cite{bedk:hf}. 

At IBM Research, we employ HF optimization techniques for sequence training \cite{Martens2010}. One of the drawbacks of this method is that training can be very slow, requiring about 3 weeks for training a 300-hour Switchboard task \cite{bedk:hf} using 64 parallel machines. There are two reasons why training is slow. Firstly, a great number of Krylov subspace iterations may be required for a solution to approximate the Hessian within each HF iteration \cite{Martens2010}, \cite{bedk:hf}. Secondly, \cite{bedk:hf} proposes using a fixed amount of data for all HF iterations in both the gradient and Krylov subspace iteration computations. The purpose of this research is to explore algorithmic strategies for reduction of the amount of time spent in both gradient and Krylov subspace computations, both by reducing the amount of data needed for training, as well as by reducing the number of Krylov subspace iterations.

In this paper, we exploit a specific instance of Krylov subspace solvers which are consumed to symmetric positive definite matrices, known as conjugate gradient (CG) solvers. For simplicity, we will use the term ``conjugate gradient"  as the specific Krylov subspace technique used to estimate the Hessian. However, the algorithms we propose for reducing training time are generic and can work with any other flexible Krylov subspace solver variant.

Preconditioning in the context of linear algebra refers to the process of transforming a system of equations into one that can be solved more readily \cite{templates1994}. For example, preconditioning has been extensively used to reduce CG iterations \cite{eisenstat:preconditioning}. Obtaining an appropriate preconditioner for a given problem can be challenging. First, the type of preconditioner that works best is problem specific. Second, while in principle, it is possible to design preconditioning strategies that will reduce the computational burden of the consequent solution phase radically, the computational investment in attaining such a preconditioner might offset its benefit. Thus, it is critical to identify a proper balance between computational efforts invested in preconditioning, vs. that invested in the consequent solution phase. 


For our optimization problem, it is computationally intractable to construct the Hessian explicitly. Quasi-Newton approaches construct (typically a low rank) an approximation to the Hessian, and in their limited memory versions, only form such approximations implicitly. In this work, we propose using the quasi-Newton L-BFGS method \cite{nocedal:lbfgs} as a preconditioner to the CG solver.  Our rationale is that while both quasi-Newton approaches and CG exploit the underlying structure of the linear(ized) system, the postulated structural assumptions of both (low rank, CG) are complementary. Therefore a combination of both methods is typically more effective than dependence upon each one solely.  The reason L-BFGS was not used directly for HF optimization of DNNs is that L-BFGS crudely approximates the curvature matrix, whereas the HF method in \cite{Martens2010} makes implicitly available the exact curvature matrix, which allows for the identification of directions with extremely low curvature.

The use of L-BFGS for preconditioning has been suggested before \cite{morales:lbfgs} for numerical simulations. We extend upon the work in \cite{morales:lbfgs}, and demonstrate that L-BFGS serves as an effective preconditioner for CG-based HF training of DNNs on large-scale speech recognition data. Furthermore, unlike \cite{morales:lbfgs} which used a typical fixed CG approach, we make here an important observation that non-fixed point preconditioners, as the proposed L-BFGS, cannot be used stablely with standard CG iterative schemes \cite{ templates1994}. Thus, to ensure, stable and predictable convergence, we propose here the use of flexible variants of CG methods \cite{notay:flexCG}. These variants avoid the failures and breakdowns that their standard counterparts are susceptible to.
 
Second, we introduce a sampling strategy in which the amount of data used for gradient and CG calculations, is gradually increased. In optimization problems, gradient-based methods typically operate within two popular regimes \cite{Byrd2012}. Stochastic approximation methods (i.e. such as stochastic gradient descent) select a small sample size to estimate the gradient. These methods often decrease the objective function loss quickly during initial training iterations, albeit, during later iterations the movement of the objective function is slow. On the other end of the spectrum, sample approximation techniques compute the gradient on a large sample of data. While this computation is expensive, the gradient estimates are much more reliable, and the objective function progresses well during later training iterations. In this study, we propose a hybrid method that captures the benefits of both stochastic and sample approximation methods, by increasing the amount of sampled data used for gradient and CG calculations. 

Sampling the amount of data used for gradient and CG calculations was explored in \cite{Byrd2012}, which observed the variance of the batch gradient to determine the amount of data to use for gradient and CG calculations. Alternatively, \cite{FriedlanderSchmidt2012} explored geometrically increasing the amount of data used for logistic regression and conditional random field problems. The benefit of this approach is that the schedule for selecting data is given ahead of time, and there is no need to compute the expensive gradient variance. In this paper, we extend the idea in \cite{FriedlanderSchmidt2012} for HF DNN training, and compare this to the sampling approach in \cite{Byrd2012}.

Initial experiments are conducted on a 50-hr English Broadcast News (BN) task \cite{bedk:hf}. We find that preconditioning allows for more than 20\% speedup by reducing the number of CG iterations. Furthermore, we find that gradient and CG sampling provide roughly additional 20\% improvement in training time. In total, by combining both sampling and preconditioning speedup ideas we were able to reduce overall training time by a factor of 1.5. Second, we extend the preconditioning and sampling ideas to a larger 300-hr Switchboard (SWB) task, where we find that the proposed techniques provide more than a 2.3x speedup, with no loss in accuracy.

\section{Hessian-Free Optimization}
Before describing the speedups made to the Hessian-free (HF) algorithm, we briefly summarize the HF algorithm for DNN training, as described in \cite{Martens2010}. Let $\boldsymbol{\theta}$
denote the network parameters, $\mathcal{L}(\boldsymbol{\theta})$
denote a loss function, $\nabla \mathcal{L}(\boldsymbol{\theta})$
denote the gradient of the loss with respect to the parameters,
$\mathbf{d}$ denote a search direction, and
$\mathbf{B(\boldsymbol{\theta})}$ denote a matrix characterizing the
curvature of the loss around $\boldsymbol{\theta}$ (i.e., a Hessian approximation). 
The central idea
in HF optimization is to iteratively form a quadratic
approximation to the loss,
\begin{equation}
\mathcal{L}(\boldsymbol{\theta}+\mathbf{d}) \approx
\mathcal{L}(\boldsymbol{\theta}) + \nabla
\mathcal{L}(\boldsymbol{\theta})^{T} \mathbf{d} + \frac{1}{2}
\mathbf{d}^{T} \mathbf{B(\boldsymbol{\theta})} \mathbf{d}
\end{equation}
and to minimize this approximation using Krylov subspace methods, such as conjugate gradient (CG),
which access the curvature matrix only implicitly through matrix-vector
products of the form $\mathbf{B(\boldsymbol{\theta})} \mathbf{d}$. Such products can be
computed efficiently for neural networks~\cite{Pearlmutter1994}. In the HF algorithm, the CG
search is truncated, based upon the relative improvement in the approximate
loss. The curvature matrix is often chosen to be the
Gauss-Newton matrix $\mathbf{G(\boldsymbol{\theta})}$~\cite{Schraudolph2002}, which may not be positive definite. 
To avoid breakdown of CG due to negative curvature, a positive definite approximation can be enforced by shifting the matrix using an additional damping term:
$\mathbf{B(\boldsymbol{\theta})} = \mathbf{G(\boldsymbol{\theta})} + \lambda \mathbf{I}$, 
where $\lambda$ is set via the Levenberg-Marquardt algorithm. 
\begin{algorithm}
\caption{Hessian-free optimization (after~\cite{Martens2010}).}
\label{alg:HF}
\begin{algorithmic}
\State initialize $\boldsymbol{\theta}$;
$\mathbf{d}_{0} \gets \mathbf{0}$;
$\lambda \gets \lambda_{0}$;
$\mathcal{L}_{\text{prev}} \gets \mathcal{L}(\boldsymbol{\theta})$
\While{not converged}
\State $\mathbf{g} \gets \nabla \mathcal{L}(\boldsymbol{\theta})$
\State Let $q_{\boldsymbol{\theta}}(\mathbf{d}) =
\nabla \mathcal{L}(\boldsymbol{\theta})^{T} \mathbf{d} +
\frac{1}{2} \mathbf{d}^{T}
(\mathbf{G(\boldsymbol{\theta})} + \lambda \mathbf{I})
\mathbf{d}$
\State $\{\mathbf{d}_{1}, \mathbf{d}_{2}, \dotsc , \mathbf{d}_{N}\}
\gets
\Call{CG-Minimize}{q_{\boldsymbol{\theta}}(\mathbf{d}),
\mathbf{d}_{0}}$
\State $\mathcal{L}_{\text{best}} \gets
\mathcal{L}(\boldsymbol{\theta} + \mathbf{d}_{N})$
\For{$i \gets N-1, N-2, \dotsc , 1$}
\Comment {{\em line search}}
\State $\mathcal{L}_{\text{curr}} \gets
\mathcal{L}(\boldsymbol{\theta} + \mathbf{d}_{i})$
\If{$\mathcal{L}_{\text{prev}} \geq \mathcal{L}_{\text{best}}
\wedge \mathcal{L}_{\text{curr}} \geq \mathcal{L}_{\text{best}}$}
\State $i \gets i + 1$
\State {\bf break}
\EndIf
\State $\mathcal{L}_{\text{best}} \gets \mathcal{L}_{\text{curr}}$
\EndFor
\If{$\mathcal{L}_{\text{prev}} < \mathcal{L}_{\text{best}}$}
\State $\lambda \gets \frac{3}{2} \lambda$; $\mathbf{d}_{0} \gets \mathbf{0}$
\State {\bf continue}
\EndIf
\State $\rho =
(\mathcal{L}_{\text{prev}} - \mathcal{L}_{\text{best}}) /
q_{\boldsymbol{\theta}}(\mathbf{d}_{N})$
\If{$\rho < 0.25$}
\State $\lambda \gets \frac{2}{3} \lambda$
\ElsIf{$\rho > 0.75$}
\State $\lambda \gets \frac{3}{2} \lambda$
\EndIf
\State $\boldsymbol{\theta} \gets \boldsymbol{\theta} + \alpha
\mathbf{d}_{i}$;
$\mathbf{d}_{0} \gets \beta \mathbf{d}_{N}$;
$\mathcal{L}_{\text{prev}} \gets \mathcal{L}_{\text{best}}$
\EndWhile
\end{algorithmic}
\end{algorithm}
Our implementation of HF optimization, which is illustrated
as pseudo-code in Algorithm~\ref{alg:HF}, closely follows that
of~\cite{Martens2010}. Gradients are computed over all the training data.
Gauss-Newton matrix-vector products are computed over a sample
(about 1\% of the training data) that is taken each time {\tt
CG-Minimize} is called. The loss,
$\mathcal{L}(\boldsymbol{\theta})$, is computed over a held-out set.
{\tt CG-Minimize}$(q_{\boldsymbol{\theta}}(\mathbf{d}),\mathbf{d}_{0})$
uses CG to minimize
$q_{\boldsymbol{\theta}}(\mathbf{d})$, starting with search direction
$\mathbf{d}_{0}$. This function returns a series of steps
$\{\mathbf{d}_{1}, \mathbf{d}_{2}, \dotsc , \mathbf{d}_{N}\}$ that are
then used in a line search procedure. The parameter update,
$\boldsymbol{\theta} \gets \boldsymbol{\theta} + \alpha
\mathbf{d}_{i}$, is based on an Armijo rule backtracking line search.
Distributed computation to computer gradients and curvature matrix-vector products is done using a master/worker architecture \cite{bedk:hf}.

\section{Preconditioning}

\begin{figure}[h!]
\centering
  \includegraphics[width=3 in]{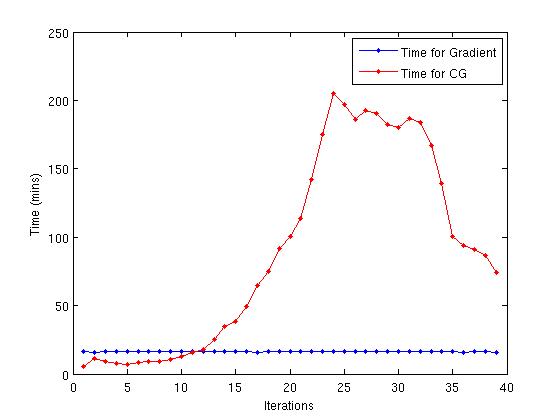}\\
  \vspace{-0.1in}
  \caption{Time spent in gradient and CG per HF iteration}
  \label{fig:baseGradCG}
  \vspace{-0.2in}
\end{figure}

One of the problems with the HF technique used in \cite{bedk:hf} is that CG algorithms used to obtain an approximate solution to the Hessian require many iterations. Figure \ref{fig:baseGradCG} indicates that as HF training iterations increase, training time per iteration is in fact dominated by CG iterations. In this section, we discuss how to reduce the number of CG iterations using preconditioning.

\subsection{Motivation}
2nd-order optimization techniques require computation of Hessian in order to determine a search direction of the  form $d_k = -H_k^{-1} g_k$. In this formulation, $H_k$ is the Hessian approximation and $g_k$ the gradient of the objective function at the $k^{th}$ HF iteration. The aforementioned CG method can be used to solve for this search direction. Specifically, we set $H_k = (G_k+\lambda I)$, where $G_k$ is the Gauss-Newton matrix, and solve $H_k d_k = -g_k$.

As mentioned above, in principle, L-BFGS \cite{nocedal:lbfgs} can be used for optimization of the HF DNN training problem. The reason L-BFGS was not used for optimization of neural networks is that in practice L-BFGS crudely approximates curvature of such systems, whereas for this domain problem HF algorithms manage to capture salient features of the curvature, and thereby identify search directions of extremely low curvature \cite{Martens2010}. 

Yet, the computation of each HF search direction can be computationally excessive, requiring a great number of CG iterations. Thus, the use of quasi-Newton methods for preconditioning such implicit systems is sensible, as the structural assumptions of CG and L-BFGS are complementary. In the section below, we describe the L-BFGS algorithm and detail using this as a preconditioner for flexible CG.


\subsubsection{L-BFGS algorithm}
L-BFGS is a quasi-Netwton optimization method that uses a limited memory technique to approximate the Hessian or its inverse. Specifically, instead of computing the Hessian directly, which can often be a large and dense matrix, the L-BFGS algorithm stores a small number of vectors which can be used as a low rank approximation of the Hessian. The L-BFGS algorithm is outlined below in Algorithm \ref{alg:LBFGS}.

\begin{algorithm}
\caption{L-BFGS Algorithm}
\label{alg:LBFGS}
\begin{algorithmic}
  \State Position at iteration $k$: $x_k$
  \State $g_k = \Delta  f(x_k)$, where $f$ is the function to be minimized
  \State $s_k = x_{k+1} - x_{k}$
  \State $y_k = g_{k+1} - g_k$
  \State $\rho_k = \frac{1}{y^T_k s_k}$
  \State Initial Hessian: $H^0_k = \frac{y^T_k s_k}{y^T_k y_k} I$
  \State $q = g_k$
  \For{$i \gets k-1, k-2, \dotsc, k-m$}
  \State $\alpha_i = \rho_i s_i^T q$
  \State $q = q - \alpha_i y_i$
  \EndFor
  \State $z=H^0_kq$
  \For{$i \gets k-m, k-m+1, \dotsc, k-1$}
  \State $\beta_i = \rho_i y_i^T z$
  \State $z = z + s_i(\alpha_i - \beta_i)$
  \EndFor
  \State $H_kg_k = z$  \Comment{search direction}
\end{algorithmic}
\end{algorithm}
\vspace{-0.2in}

\subsubsection{L-BFGS as a Preconditioner}

CG iterative methods can be used to solve for the search direction $d_k$, by minimizing the following problem $H_k^{-1} g_k-d_k=0$. Preconditioning typically involves a process or transformation (e.g. change of coordinates) applied upon a system of equations, which in return, converts the system to of more favorable structure. Preconditioning makes the CG problem easier to solve and reduces the number of CG iterations. If we define $M$ as a preconditioner, preconditioned CG involves the following transformation to the CG problem $M^{-1}(H_k^{-1} g_k-d_k)$. The preconditioner $M$ is required to be symmetric and positive definite, and fixed for all iterations. If any of these conditions are violated, the CG method may fail.


Prescription of a suitable preconditioning scheme for a given problem is challenging. First, each system has its own characteristic structure. Identification of which and respectively determining the type of preconditioner that works best is generally problem specific. Second, if the preconditioner is computationally expensive to obtain, then this will offset any reduction in CG iterations, and thus the preconditioner will not be cost effective. Third, as challenging as preconditioning is in ordinary circumstances, a greater challenge is to precondition an implicit system, that cannot be accessed directly.

Previous preconditioning work for HF optimization has focused on diagonal matrix preconditioners. \cite{Martens2010} explored using the diaogonal elements of the Fisher information matrix as a preconditioner for HF training of DNNs. Using diagonal matrix elements has a very limited ability to precondition the system, and is mainly beneficial when the matrix suffers scaling issues. In addition, \cite{chapelle:jacobi} explored using the Jacobi pre-conditioner, which is computed over a batch of data just like the curvature-vector products, thus requiring the master/worker data-parallelization architecture. For our specific DNN speech problem, we found that the Jacobi preconditioner was costly to compute and offset reductions in CG iterations.  The L-BFGS \cite{morales:lbfgs} preconditioner we propose is far more powerful compared to diagonal matrix preconditioners as it improves the spectral properties of the system, rather than merely tackling potential scaling issues. Furthermore, it does not require any data parallelization.

The L-BFGS preconditioner is described as follows. Each iteration $i$ of CG produces a sequence of iterates $x_i$ (i.e., $d_i$ in Algorithm 1) and a sequence of residuals $r_i$ \cite{shewchuk1994introduction}. Using these statistics, the vectors $s_i = x_{i+1} - x_i$ and $y_i = r_{i+1} - r_i$ are stored for $m$ iterations of CG, where $m$ is specified by the user. Once $m$ statistics are saved, an L-BFGS matrix $H$ can be defined using the steps in Algorithm \ref{alg:LBFGS}. This L-BFGS matrix is used as the preconditioner for CG.

There are a variety of different methodologies to choose the $m$ statistics to use when estimating the L-BFGS matrix. We adopt a strategy proposed in  \cite{morales:lbfgs}, namely using $m$ vectors evenly distributed throughout the CG run, to estimate the L-BFGS matrix. This implies that our preconditioner changes for different CG iterations. The requirement that 
the preconditioner needs to be fixed for all iterations of CG is inconvenient, since as we obtain more L-BFGS statistics we can improve the estimate of the preconditioner. 
Changing the preconditioner for CG requires using a flexible CG approach \cite{notay:flexCG}.  
More specifically, instead of using the equivalent of Fletcher-Reeves updating formula for non-preconditioned CG, the Polak-Ribi\`{e}re variant is required \cite{shewchuk1994introduction}. 
This is opposed to the approach taken in \cite{morales:lbfgs} which did not use a flexible CG approach.

\section{Sampling}

Another problem with the HF technique used in \cite{bedk:hf} was that the gradient was computed using all data, and CG on a fixed data sample. In this section, we explore reducing the amount of data used for the gradient and CG computations. 
Specifically, we explore a hybrid technique that first starts with a small amount of data similar to stochastic approximation methods, and gradually increases the amount of sampled data similar to sample approximation methods. In the following section, we detail two different hybrid methods. 

\subsection{Sampling From Variance Estimates} 

\cite{Byrd2012} proposes a method to increase the sample size based on variance estimates obtained during the computation of a batch gradient.  This algorithm can be described as follows. Denote $f(w;x_i)$ as the output from the DNN and $y_i$ the true output, such that a loss between predicted and true values can be defined as $l(f(w;x_i), y_i)$. The loss over the training set of size $N$, is defined as the sum of the losses from the individual training examples $x_i$, as shown by Equation \ref{eq:gradTotal}.

\begin{equation}
J(w) = \frac{1}{N}  \sum_{i=1}^N l(f(w; x_i), y_i)
\label{eq:gradTotal}
\end{equation}

In addition, the loss over a subset $S \subset \{1, \ldots, N\}$  is defined by Equation \ref{eq:gradSubset}. 

\begin{equation}
J_S(w) = \frac{1}{S}  \sum_{i \subset S} l(f(w; x_i), y_i)
\label{eq:gradSubset}
\end{equation}

Denoting the gradients of the full and subset losses as $\nabla J(w)$ and $\nabla J_S(w)$ respectively, the algorithm ensures that descent made in $J_S$ at every iteration must admit a descent direction for the true objective function $J$. The is expressed by Equation \ref{eq:gradRelation}. 

\begin{equation}
\delta_S(w) \equiv || \nabla J_S(w) - \nabla J(w) ||_2 \leq \theta ||\nabla J_S(w) ||_2, \text{ where } \theta \in [0,1)
\label{eq:gradRelation}
\end{equation}

In practice, the quantity $\delta_S(w)$ is not evaluated (the computation of $\nabla J(w)$ is expensive for large data sets),  
but instead is estimated from the variance of $\nabla J_S(w)$. Inequality \ref{eq:gradRelation} can be simplified to the inequality 
\begin{equation}
\frac{||Var_{i \in S}(\nabla l(w;i)) ||_1}{|S|}  \leq \theta^2 ||\nabla J_S(w) ||_2^2\;.
\label{eq:gradRelationAgain}
\end{equation}
If this inequality fails, a new sample size $\hat S > S$ is selected to satisfy Inequality \ref{eq:gradRelationAgain}. 
The same dynamic selection strategy is also applied to the CG iterations. 

In this paper, we explore this sampling approach within a DNN framework. Given an input utterance $u$, the output of the DNN is the sum of the gradients of all training frames $L$ in that utterance, i.e. $\sum_{i=1}^L \nabla l(w;i)$. Therefore, to compute the variance of the gradient estimate, this requires two passes through each utterance to compute the gradient and gradient-squared statistics $\sum_{i=1}^L \nabla l^2(w;i)$. Since this makes the algorithm computationally expensive, we compute the average gradient per utterance $u$, i.e. $\bar{l}_u = \frac{1}{L}\sum_{i=1}^L \nabla l(w;i)$. The variance statistics become the sum and sum-squared of $\bar{l}_u$ over all utterances $u \in S$ in the training set, as shown by Equation~\ref{eq:estVar}. This only requires one pass through the network per utterance.

\begin{equation}
Var_{i \in S} (\nabla l(w;i)) \approx \frac{\sum_{u=1}^S \bar{l}_u^2 - (\sum_{u=1}^S \bar{l}_u )^2/S}{S-1}
\label{eq:estVar}
\end{equation}

\subsection{Geometric Sampling}

The sampling approach proposed above uses sampling statistics to approximate 
the descent condition~\eqref{eq:gradRelationAgain}, 
but the need to estimate the variance in~\eqref{eq:gradRelationAgain} adds
notable computational complexity to the gradient computation. 
In contrast, the framework discussed in~\cite{FriedlanderSchmidt2012} 
provides an expected guarantee of descent in each iteration, as long as 
the sampling errors  
\[
E [|| \nabla J_S(w) - \nabla J(W) ||_2^2] \leq B_k
\]
are bounded, and the bounds $B_k$ are decreasing. In fact,~\cite[Theorem 2.2]{FriedlanderSchmidt2012}
links the sampling errors directly to the expected rate of convergence. 
This approach does not require computing statistics along the way, 
and the sampling strategy used to select $S$ is linked 
directly to the expected convergence rate in~\cite{FriedlanderSchmidt2012,AravkinFHV:2012}.

~\cite{FriedlanderSchmidt2012} uses a geometrically increasing sample size. 
We adopt this strategy for the gradient and CG iteration samples in each iteration. 
Specifically, given initial sample size $S_0$, the sample size at each iteration $i$ is given by Equation \ref{eq:geom} where $\alpha$ is the geometric factor that is tuned on a development set.

\begin{equation}
|S_i| = \alpha^i |S_0|
\label{eq:geom}
\end{equation}

This approach fits into the theory proposed in~\cite{FriedlanderSchmidt2012}, and has the practical benefit of 
{\it a priori} sample size selection. 
The sample size can be used both for gradient and CG iteration calculations.
\section{Experiments \label{sec:experiments}}

\subsection{Broadcast News}

Our initial experiments are conducted on a 50-hr English Broadcast News (BN) task and results reported on both the EARS {\tt dev04f} set. We use a recipe outlined in  \cite{Soltau2010} to extract acoustic features. The hybrid DNN is trained using speaker-adapted VTLN+fMLLR features as input, with a context of 9 frames around the current frame. In \cite{bedk:hf}, it was observed that a 5-layer DBN with 1,024 hidden units per layer and a sixth softmax layer with 2,220 output targets was an appropriate architecture for BN tasks. 

We explore the behavior of preconditioning and sampling for HF training on a smaller BN task first, before moving to a larger Switchboard task. All timing experiments in this study were run on an 8 core Intel Xeon X5570@2.93GHz CPU. Matrix/vector operations for DNN training are multi-threaded using Intel MKL-BLAS. 12 machines were exclusively reserved for HF training to get reliable training time estimates.


\section{Results \label{sec:results}}

\subsection{Preconditioning}
In this section, we compare CG with preconditioning and no preconditioning (noPC). For preconditioning, we explore the behavior with different number of statistics used to estimate the L-BFGS preconditioned, namely 16 (PC-16), 32 (PC-32) and 64 (PC-64).

\begin{figure}[h!]
\centering
  \includegraphics[width=3 in]{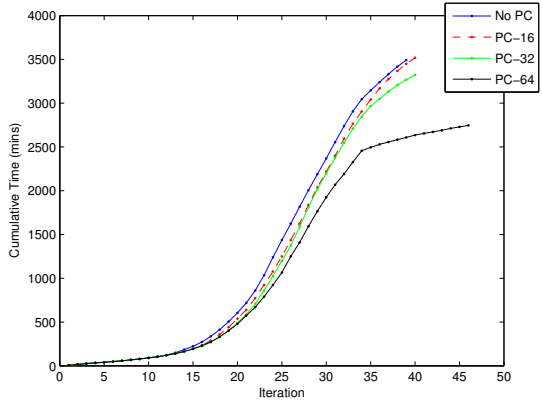}\\
  \vspace{-0.1in}
  \caption{Cumulative CG runtime for different PC schemes}\label{fig:pcTime}
  \vspace{-0.1in}
\end{figure}

Table \ref{table:CGtime} shows the total time spent in CG, and total number of training iterations, to achieve the same loss. In addition, Figure \ref{fig:pcTime} provides a closer look at the cumulative time for CG for the 4 methods. The Figure indicates that that all preconditioning methods require less time for CG, particularly as the number of total HF iterations increases (and thus the number of CG iterations increases). We see that PC-64 manifests a significant reduction in CG time after 30 HF iterations, but this also results in the loss moving much slower for this method, as explained by increased HF iterations in Table \ref{table:CGtime}. PC-32 appears to be the most cost-efficient choice for the given task, both in terms of CG iteration runtime and in terms of loss reduction, and is roughly 22\% faster than the baseline method. 

\begin{table} [h!]
  \centering
  \begin{tabular}{|c||c|c|c|}
    \hline
    Method & Loss & HF Iterations & Time (min) \\ \hline
    noPC &  1.9153 & 39 &  3,492.2 \\ \hline
    PC-16 &   1.9157 & 35 &  3,042.2\\ \hline
    PC-32 & 1.9150 & 33 &  2,7095.3\\ \hline
    PC-64 &   1.9158 & 46 &  2,745.6 \\ \hline
  \end{tabular}
   \caption{Total CG runtime for different quasi-Newton PC schemes}\label{table:CGtime}
  \vspace{-0.1in}
\end{table}

%

\subsection{Gradient+CG Sampling}

Next, we compare the behavior of the geometric and variance sampling methods. Sampling methods require a tradeoff between amount of data used, and the number of iterations for the training loss to converge. Using too little data for gradient and CG will require more training iterations, while using too much data will make each iteration computationally expensive. 

For geometric sampling, the geometric factor $\alpha$ was tuned on a held-out set for both gradient and CG. It was found that an $\alpha_g=1.2$ for the gradient and $\alpha_{cg}=1.3$ for CG allowed for the best tradeoff between reduction in amount of training data used and training time.  This geometric factor corresponds to seeing roughly 100\% of the total data used for gradient and CG calculations when roughly 50\% of the total training iterations are completed. For variance sampling, $\theta$ in Equation \ref{eq:estVar} is tuned, where smaller $\theta$ favors a larger sample size. 

\begin{figure}[h!]
\centering
  \includegraphics[width=3 in]{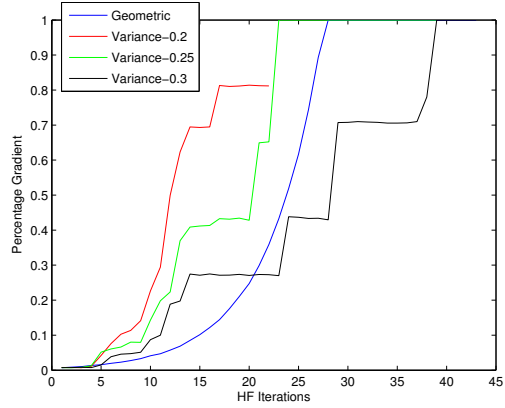}\\
  \vspace{-0.1in}
  \caption{Percentage of Gradient Accessed for Sampling Methods}\label{fig:gradSampling}
\vspace{-0.1in}
\end{figure}

Figure \ref{fig:gradSampling} shows the percentage of data accessed for the gradient for the geometric and variance methods, per HF iteration, for three different values of $\theta$. Notice that the variance methods access a lot of training data in the beginning relative to the geometric method. One reason is that during the beginning of training, there is little data available to get a reliable variance estimate, so a larger sample size is preferred. The variance method with $\theta=0.25$ provided the best tradeoff between training time and data accessed. A similar $\theta$ was also used for estimating amount of data used for CG. 

\begin{figure}[h!]
\centering
  \includegraphics[width=3 in]{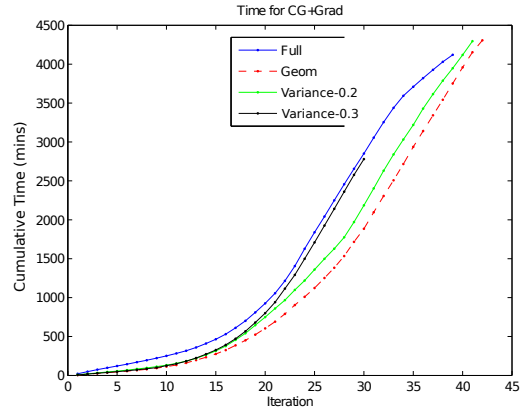}\\
  \vspace{-0.1in}
  \caption{Cumulative Training Time for Sampling Methods}\label{fig:samplingTime}
\vspace{-0.1in}
\end{figure}

Figure~\ref{fig:samplingTime} shows the cumulative time for gradient and CG calculation per HF iteration, for the full gradient/CG and sampling approaches, where both sampling approaches are tuned to provide best tradeoff between training time and amount of data accessed. The geometric method is quicker than the variance sampling method, particularly because it accesses less data during early training iterations, as shown in Figure \ref{fig:gradSampling}. Overall, we find that the geometric method provides roughly a 20\% reduction in training time. It is possible that a technique that starts with geometric sampling, and then switches to variance sampling once enough data is obtained for a reliable variance estimate, might provide further speedups.

\subsection{Overall Speedups}

\begin{figure}[h!]
\centering
  \includegraphics[width=3 in]{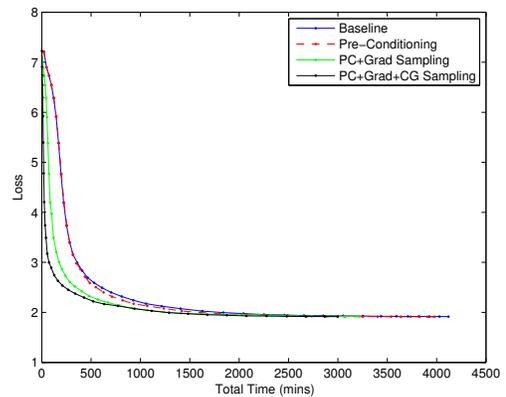}\\
  \caption{Loss vs. Training Time for Different Speedup Techniques}\label{fig:timingAll}
\end{figure}

In this section, we combine the preconditioning and sampling to calculate the overall speedup in training time for BN. Figure \ref{fig:timingAll} shows the trade-off between loss and overall training time of the baseline (no speedup) method, preconditioning, and then including gradient and CG sampling. Overall we can see that PC+Gradient+CG sampling offers the fastest training time compared to the baseline. Table \ref{table:bnSpeedups} shows the training time and corresponding WER for the baseline and speedup methods. Training time is reduced from 68.7 hours to 44.5 hours, roughly a 1.5x speedup, with no loss in accuracy.

\begin{table}[h!]
\begin{center}
\begin{tabular}{|c|c|c|} \hline
Method & WER & Total Training Time (hrs) \\ \hline
Baseline & 17.8 & 68.7 \\ \hline
PC+Grad+CG Speedups & 17.8 & 44.5 \\ \hline
\end{tabular}
\end{center}
\vspace{-0.1in}
\caption{Overall Training Time Improvements, Broadcast News}
\label{table:bnSpeedups}
\vspace{-0.2in}
\end{table}%

\subsection{Speedups on Larger Task}

After analyzing the behavior of preconditioning and sampling speedups on a smaller 50-hour Broadcast News task, in this section, we explore training speed improvements on a larger 300-hour Switchboard task. 
\subsubsection{Experimental Setup}

We explore DNNs performance on 300 hours of conversational American English telephony data from the Switchboard corpus.  Development is done on the {\tt  Hub5'00} set, while testing is done on the {\tt rt03} set, where we report performance separately on the Switchboard ({\tt SWB}) and Fisher ({\tt FSH}) portions of the set.

Similar to BN, the training features are speaker-adapted, using VTLN and fMLLR techniques. The input features into the DNN have an 11-frame context  ($\pm5$) around the current frame. Similar to \cite{bedk:hf}, the DNN has six hidden layers each containing 2,048 sigmoidal units, and 8,260 output targets. Results with and without HF speedups are reported after sequence training. 

\subsubsection{Results}

Performance with the baseline and speedup HF techniques are shown in Table \ref{table:swbSpeedups}. Since using 32 L-BFGS stats performed well for the smaller 50-hour BN task, we used the same on the Switchboard task for preconditioning. In addition, because of the increased amount of training data associated with the larger task, we found that using a smaller sample size (i.e., $\alpha$) for the gradient and CG iteration calculations still allowed for an appropriate estimate of these statistics.

Since we use more parallel machines (i.e. 64) for SWB compared to BN, it was not possible to exclusively reserve machines for timing calculations. Therefore, training time is estimated by calculating total number of accessed data points for training, which is correlated to timing. Table \ref{table:swbSpeedups} shows the total accessed data points for the baseline and speedup techniques. Notice that with a larger dataset, because we are able to decrease the fraction of data used for gradient and conjugate gradient calculations, was can achieve an even larger speedup of 2.3x over the baseline, with no loss in accuracy. This suggests that even further speedups are possible as the data size grows.

\begin{table}[h!]
\begin{center} 
\begin{tabular}{|c|c|c|} \hline
 Method & WER & Total Accessed Data Points\\ \hline
Baseline & 12.5 & 2.26e9 \\ \hline
PC+Grad+CG Speedups & 12.5 & 9.95e8 \\ \hline
\end{tabular}
\end{center}
\vspace{-0.1in}
\caption{Overall Training Time Improvements, Switchboard}
\label{table:swbSpeedups}
\vspace{-0.1in}
\end{table}%

\section{Conclusions \label{sec:conclusions}}

In this paper, we explored using an L-BFGS pre-conditioner and geometric sampling approach to accelerate HF training. We find that both approaches combined provided roughly a 1.5x speedup over a 50-hr Broadcast News task and a 2.3x speedup on a 300-hr Switchboard task, with no loss in accuracy. We anticipate an even larger speedup to be attained by more informed selection of quasi-Newton statistics (potentially adaptive) as well as by application of the proposed algorithmic strategies upon problems of greater scale. 

\bibliographystyle{IEEEbib}
\begin{footnotesize}
\bibliography{main}
\end{footnotesize}
\end{document}